\documentclass[letterpaper, 10 pt, journal, twoside]{IEEEtran}

\ifCLASSINFOpdf
\else
\fi
\usepackage{graphicx}
\usepackage{booktabs}
\usepackage[table,xcdraw]{xcolor}
\usepackage{multirow} 
\usepackage{amsmath}
\usepackage{amssymb}
\usepackage{epsfig}
\usepackage{hyperref}
\usepackage{cleveref}
\usepackage{mathrsfs}
\newcommand\Tstrut{\rule{0pt}{2.6ex}}         
\newcommand\Bstrut{\rule[-0.9ex]{0pt}{0pt}}   
\begin{document}

\title{Hybrid Attention for Robust RGB-T Pedestrian Detection in Real-World Conditions}

\author{Arunkumar Rathinam$^{1}$$^{\ast}$, Leo Pauly$^{2}$$^{*}$, Abd El Rahman Shabayek$^{1}$, Wassim Rharbaoui$^{3}$, \\ Anis Kacem$^{1}$, Vincent Gaudillière$^{4}$ and Djamila Aouada$^{1}$ \\ ($^{*}$Equal contribution)
\thanks{Manuscript received: May, 27, 2024; Revised September, 17, 2024; \\ Accepted October, 23, 2024.}
\thanks{This paper was recommended for publication by Editor Gentiane Venture upon evaluation of Reviewers' comments.
This work was supported by Luxembourg National Research Fund (FNR), under the project reference BRIDGES2020/IS/14755859/MEET-A/Aouada.} 
\thanks{$^{1} $Authors are with Interdisciplinary Center for Security, Reliability and Trust (SnT), University of Luxembourg, L-1855 Kirchberg Luxembourg.
        {\tt\footnotesize arunkumar.rathinam@uni.lu}}%
\thanks{$^{2} $The work was done when author was with the SnT, University of Luxembourg.}
\thanks{$^{3} $The author was with the SnT, University of Luxembourg. He is now with the XLIM institute, University of Poitiers, F-87060 Limoges, France. }
\thanks{$^{4} $The author was with the SnT, University of Luxembourg. He is now with Université de Lorraine, CNRS, Inria, Loria, F-54000, Nancy, France.}
\thanks{Code is available at \href{https://cvi2.uni.lu/ha-mlpd/}{https://cvi2.uni.lu/ha-mlpd/} }
}

\markboth{IEEE Robotics and Automation Letters. Preprint Version. Accepted October, 2024}
{ \MakeLowercase{\textit{}}} 

\maketitle

\begin{abstract}
Multispectral pedestrian detection has gained significant attention in recent years, particularly in autonomous driving applications. To address the challenges posed by adversarial illumination conditions, the combination of thermal and visible images has demonstrated its advantages. However, existing fusion methods rely on the critical assumption that the RGB-Thermal (RGB-T) image pairs are fully overlapping. These assumptions often do not hold in real-world applications, where only partial overlap between images can occur due to sensors configuration. Moreover, sensor failure can cause loss of information in one modality. In this paper, we propose a novel module called the Hybrid Attention (HA) mechanism as our main contribution to mitigate performance degradation caused by partial overlap and sensor failure, \textit{i.e.} when at least part of the scene is acquired by only one sensor. We propose an improved RGB-T fusion algorithm, robust against partial overlap and sensor failure encountered during inference in real-world applications. We also leverage a mobile-friendly backbone to cope with resource constraints in embedded systems. We conducted experiments by simulating various partial overlap and sensor failure scenarios to evaluate the performance of our proposed method. The results demonstrate that our approach outperforms state-of-the-art methods, showcasing its superiority in handling real-world challenges. 
\end{abstract}

\IEEEpeerreviewmaketitle

\section{Introduction}
\IEEEPARstart{P}{edestrian} detection is one of the important domains within computer vision for robotics, playing a significant role in applications such as self-driving vehicles, surveillance automation, and mobile robot navigation~\cite{hwang2015multispectral}. RGB cameras are commonly preferred sensors for such applications. However, they tend to suffer from overexposure in daylight, low illumination in night scenarios, and high-contrast lighting. To address these shortcomings, a number of sensors and fusion solutions were investigated. In particular, thermal cameras seem to provide several advantages in terms of costs, algorithms, and data~\cite{krotosky2007color}. Among them, RGB images provide texture and color information, while thermal images focus on the infrared heat emitted by the objects and are therefore invariant to lighting conditions~\cite{GUO2021110176}. RGB and thermal images are thefore complementary with each other by nature. This led the community to collect multispectral datasets such as KAIST~\cite {hwang2015multispectral}, CVC~\cite{GonzalezFSSVXL16} or FLIR~\cite{FLIR}, providing thermal data in addition to RGB data. 

KAIST dataset provides fully-overlapping RGB-T image pairs, i.e. both images are acquired at the same time and cover the same field of view. However, acquiring such image pairs requires specialised sensor setup over conventional stereo setup which is widely used in real-world applications. In stereo setups, partial overlap will occur inherently due to a different camera Field of View (FoV) and pixel-level misalignment will occur due to parallax~\cite{kim2021mlpd}.
\begin{figure}
  \includegraphics[scale=.32]{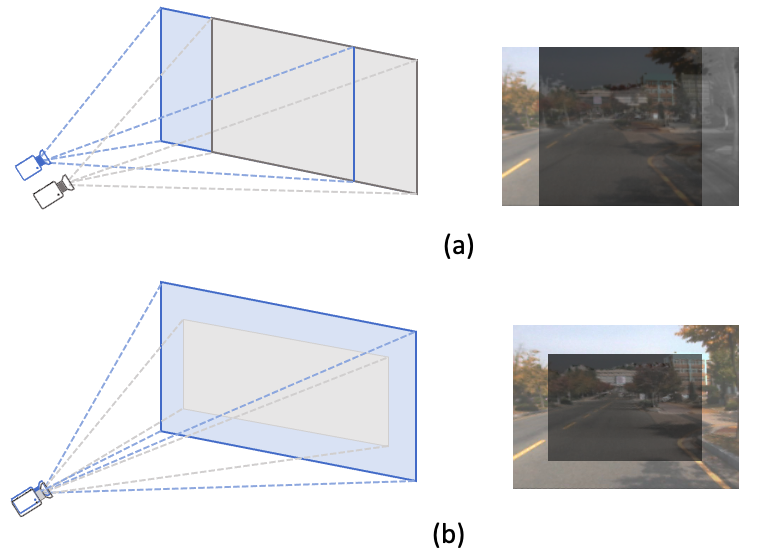}
  \caption{Constraints on (a) stereo setup (extrinsic parameters) and (b) sensor resolution (intrinsic parameters) resulting in only partial overlap between RGB and thermal images.}
  \label{fig:fig1}
  \vspace{-1.5em}
\end{figure}
Information discrepancy between one image and the other can cause features to be out of their corresponding positions, resulting in decreased algorithm performance and less accurate predictions during the inference process~\cite{ZhangZC0LL19}. Even in the KAIST dataset that has fully overlapping image pairs, the authors attempted to reduce the pixel-level misalignment problem. This was achieved by further improving the original data labels to \textit{``sanitised"} cross-modal annotations~\cite{li_2018_BMVC} and \textit{``paired"} modality-specific annotations~\cite{ZhangZC0LL19} (in this paper, different \textit{modalities} correspond to different spectral images: RGB or thermal). 

Recent methods, such as multi-label learning~\cite{kim2021mlpd}, aimed to learn more discriminative features while using semi-unpaired augmentation to generate unpaired inputs between two modalities where considering a single bounding-box label is irrelevant. Even with different learning approaches, the robustness of an algorithm is questioned when one of the modalities is unavailable or partially available; for example, when a malfunction in one of the camera sensor arrays leads to a partial or even complete loss of one modality. This situation is investigated in very few existing literature~\cite{kim2021mlpd}, and the performance drop in such scenarios is quite high even for best performing models on fully overlapping images. 

To improve performance and enhance algorithm robustness, our research examines the issue of partial overlap caused by various factors such as stereo configurations, sensor malfunctions, and others. This can result in partial or complete invisibility of regions in one modality during inference. In this paper, we adopt the term \textit{blackout} to refer to areas in the union of pictures where data are absent from one of the modalities. \Cref{fig:fig1} shows sample cases of \textit{blackout} that can arise from sensor setups, \Cref{fig:fig1}-a showing \textit{sides blackout} from different camera extrinsics in stereo setup, \Cref{fig:fig1}-b depicting \textit{surrounding blackout} arising due to difference in camera intrinsics such as sensor resolution or focal length. To achieve robustness in such scenarios, we present a hybrid attention module, which reduces performance degradation irrespective of network architecture. We also consider a much lighter backbone compared to previous works for coping with hardware resource constraints in embedded systems. Our contributions are as follows: 
\begin{itemize}
    \item We introduce the Hybrid-Attention (HA) module, which combines self-attention and cross-attention, to mitigate performance degradation arising from modality-specific blackouts;
    \item We propose an improved RGB-T fusion algorithm, named Hybrid Attention-based Multi-Label Pedestrian Detector (HA-MLPD), robust against partial overlap and sensor failure encountered in real-world scenarios, while being resource-friendly;   
    \item We provide experimental evidence that the proposed method prevents a performance drop and makes the fusion algorithm more robust and reliable irrespective of the network backbone architecture.
\end{itemize}

The remainder of the paper is structured as follows: \Cref{sec:related_works} provides an overview of the related literature. In \Cref{sec:proposed_method}, the HA-MLPD algorithm is introduced, with a detailed explanation of its implementation. Our approach is tested on the KAIST dataset~\cite{hwang2015multispectral} under various simulated blackout conditions, as discussed in \Cref{experiments_results}. Finally, \Cref{section:conclusion_and_future_works} concludes with a discussion on future research directions.

\section{Related Works}
\label{sec:related_works}

\begin{figure*}[h]
    \centering
  \includegraphics[trim=0mm 1mm 0mm 0mm, clip, width=0.9\linewidth]{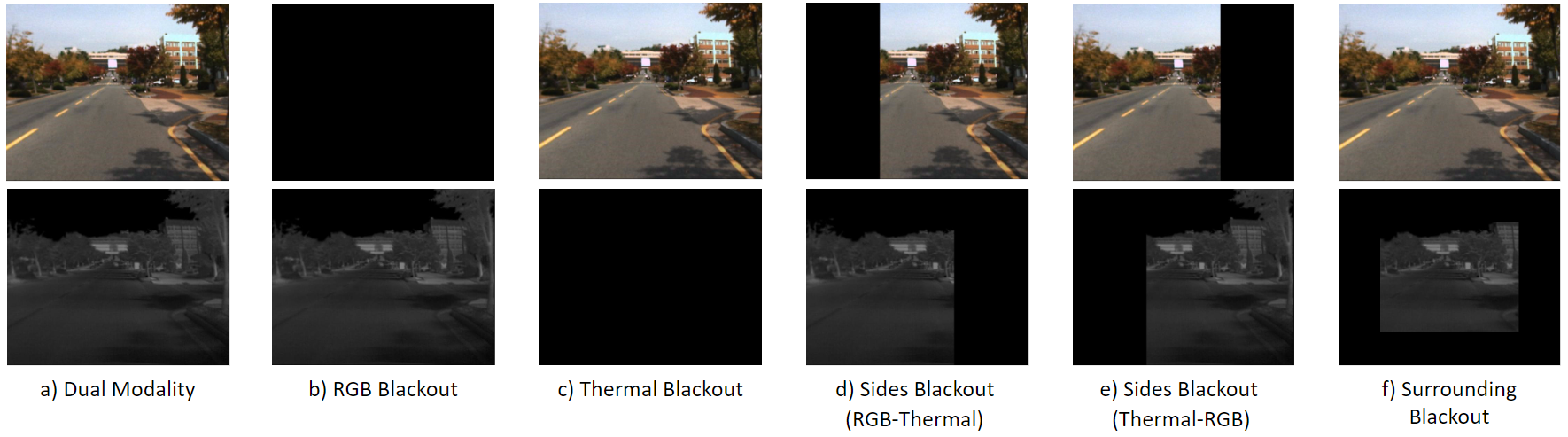}
  \vspace{-2mm}
      \caption{Simulated inference-time conditions for assessing the robustness of HA-MLPD: a) fully-overlapping RGB and thermal images; b),c) complete blackouts of one of the modalities (sensor failures); d),e),f) partial overlaps (discrepancies in the extrinsic (d,e) or intrinsic (f) camera parameters).}
  \label{fig:blackout_scenarios}
  \vspace{-1em}
\end{figure*}


The multispectral pedestrian detection problem has been widely studied in computer vision. Several classical methods use RGB and thermal images, relying on pixel difference values~\cite{lee2015robust}, local shape features~\cite{zhang2007pedestrian}, contour saliency maps~\cite{davis2005fusion}, disparity maps~\cite{krotosky2007color} and HOG features~\cite{yuan2015multi}. 

In 2015, Hwang \textit{et al.} introduced KAIST~\cite{hwang2015multispectral}, a large-scale multispectral pedestrian detection dataset that contains RGB and thermal images with the corresponding pedestrian labels. The release of the KAIST dataset accompanied a renewed interest in the multispectral pedestrian detection problem, and several new methods have been proposed since then. For example, Liu \textit{ et al.}~\cite{liu1611multispectral} proposed a deep learning-based Halfway Fusion model and presented comparative analyses with other early fusion architectures (input-level fusion) and late fusion architectures (decision-level fusion). In another study, Li \textit{et al.}~\cite{li_2018_BMVC} demonstrated that the incorporation of an additional semantic segmentation task led to enhanced performance compared to the use of a model focused solely on detection. They introduced a combined architecture that included a multispectral proposal network to generate pedestrian proposals and a subsequent multispectral classification network to distinguish pedestrian instances from challenging negatives. The authors trained the integrated network by simultaneously optimising both pedestrian detection and semantic segmentation tasks. Zheng \textit{et al.}~\cite{zheng2019gfd} introduced Gated Fusion Units (GFU) which are designed to merge feature maps from the feature extraction layers of Single Shot MultiBox Detector (SSD) at various scales~\cite{SSDDetector}.
In their studies~\cite{li2019illumination, guan2019fusion}, the authors explored the use of distinct subnetworks for individual modalities and incorporated illumination-adaptive weighting of these subnetworks to enhance the efficiency of multispectral pedestrian detection. This approach enabled the prioritization of information from the RGB modality in adequately illuminated images or from the thermal modality in low-light situations. Zhou \textit{et al.}~\cite{zhou2020improving} introduced Modality Balance Network (MBNet), which simultaneously compensated for modality imbalance problems in illumination and at the feature levels. Chen \textit{et al.}~\cite{chen2022multimodal} presented a late fusion architecture by probabilistically ensembling decisions made individually from RGB and thermal images. Zhang \textit{et al.}~\cite{zhang2021guided} presented a fusion mechanism under the guidance of the intermodal and intramodal attention modules, to learn to dynamically weigh and fuse multispectral features. Yang \textit{et al.}~\cite{fphy.2023.1121311} proposed an algorithm that uses cascaded information enhancement and fusion of cross-modal attention features, both of which rely on the attention mechanism. 

The existing research has significantly enhanced the effectiveness of combining RGB and thermal images for pedestrian detection through multispectral fusion. Nonetheless, the algorithms typically operate under the assumption that both types of images are accessible during inference, neglecting scenarios where sensor failure may occur. Furthermore, potential sensor modifications during inference such as modifications in the stereo arrangement or lens settings, compared to the training dataset acquisition, are not taken into account. Such discrepancies may result in only partially-overlapping images leading to \textit{blackout} regions (refer to \Cref{fig:fig1}). Analysis of existing algorithms shows a considerable degradation in performance under such conditions~\cite{kim2021mlpd}. However, robustness to such realistic inference-time conditions is important for real-world deployment of algorithms and has not received much attention in the field. In this direction, Zhang \textit{et al.}~\cite{zhang2019weakly} proposed a region feature alignment module (RFA), which adaptively compensates for misalignment of feature maps in both modalities.  Recently, Kim \textit{et al.} ~\cite{kim2021mlpd} introduced the MLPD algorithm, which has shown robustness to sensor failures and inference-time partial overlap. 
However, MLPD lacks interconnection between the feature extraction branches of the RGB and the thermal modalities. This could affect performance during blackout scenarios.  Similarly,~\cite{kim2022towards} used Multispectral Recalling (MSR) memory that can recall the missing features of multispectral modalities to detect pedestrians when one of the modalities is available. However, this architecture is not designed to handle both available simultaneous modalities.

Beyond the domain of multispectral pedestrian detection, the issue of partial image overlap has also been investigated in related fields, such as multi-view person tracking from multiple cameras~\cite{NARAYAN2019222}. In this context, proposed solutions range from hand-engineered methods, such as geometric transformations of images based on predefined ground plane homographies~\cite{4587539,11744085_11}, to more advanced deep learning-based approaches~\cite{MA2024102496}. However, these methods require substantial adaptation to address the specific challenges posed by multispectral pedestrian detection, particularly due to the multimodal nature of the data and differences in problem formulation.

\begin{figure*}[ht]
    \centering
    \includegraphics[width=0.96\textwidth]{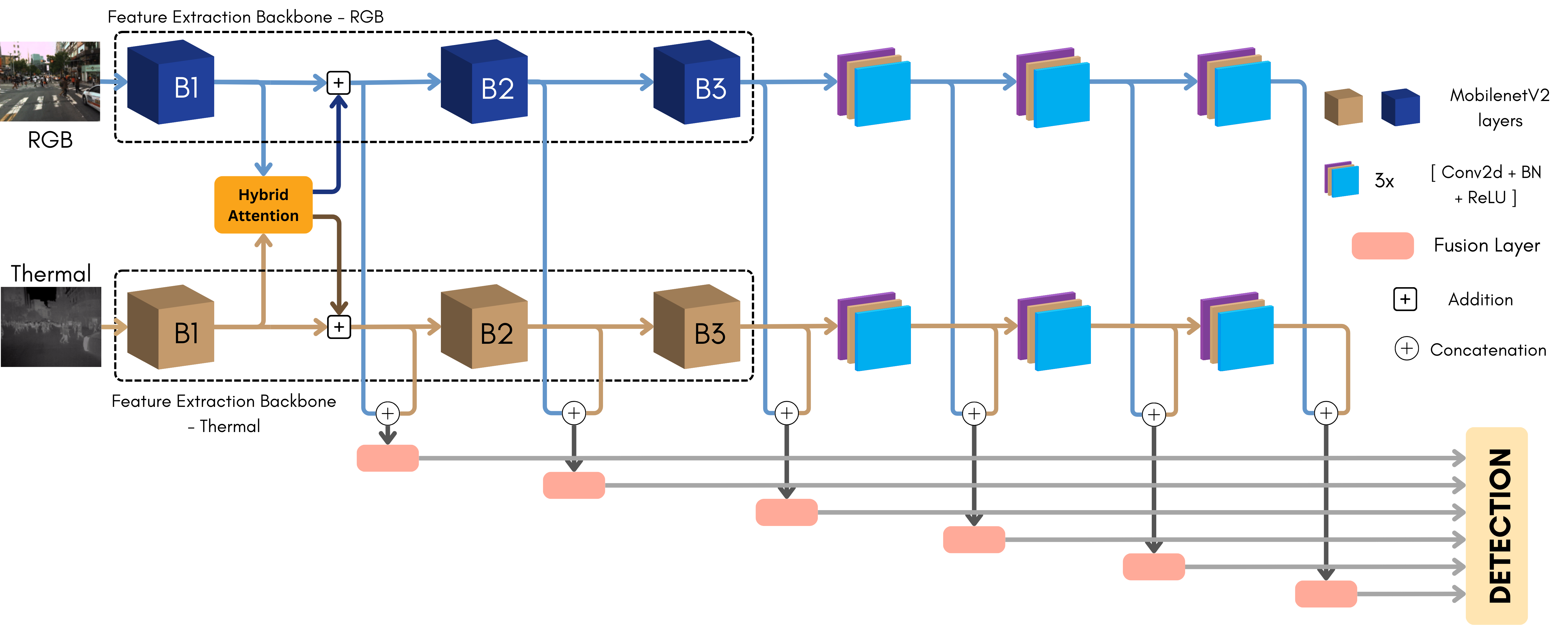}
    \vspace{-0.5em}
  \caption{Network Architecture for the proposed HA-MLPD. Modality-specific features are extracted from the respective image using a feature extraction backbone (here MobilenetV2~\cite{sandler2018mobilenetv2}, and the blocks B1, B2 and B3 are described as in \Cref{tab:mobilenetv2_architecture}). The HA module then enhances the features by attending only to the regions with useful information. This is followed by a sequence of shared convolutional blocks that extract features common to both modalities. Finally, the extracted multiscale features are fed to the SSD detector~\cite{SSDDetector} to predict the pedestrian bounding boxes and their confidence scores.}
  \label{fig:network_architecture}
\end{figure*}

In this work, we propose an RGB and thermal fusion algorithm for pedestrian detection using the novel HA module, providing robustness to blackout scenarios caused by inference-time sensor failure or partial overlap. Our proposed HA module facilitates the flow of information and interconnects the features extracted between the modalities using cross-attention~\cite{chen2021crossvit,cai2021auditory} (during normal conditions) and self-attention~\cite{vaswani2017attention,dosovitskiy2021an} (during blackouts). 

\section{Proposed HA-MLPD}
\label{sec:proposed_method}
\subsection{Problem Statement}

Our method, HA-MLPD, assumes that images from both modalities are automatically registered at test time and that pixel values in resulting non-overlapping regions are set to zero (i.e. blackout). \Cref{fig:blackout_scenarios} shows some examples. The masks of the overlapping regions are further leveraged to guide the network through cross- or self-attention to the features. For that, we use masks $M_{rgb}$, $M_{thermal}$ corresponding to blackout regions as what can be obtained by registering the images from the two modalities (using methods such as~\cite{Arar_2020_CVPR} for instance, or directly using the likely known stereo parameters). Note that the registration process is not within the scope of this paper. In detail, the masks $M_{rgb}$ and $M_{thermal}$ are set to 1 modality-specific information is available at the location of the pixel, 0 otherwise.

\begin{figure*}[t]
  \centering
  \includegraphics[width=0.75\textwidth]{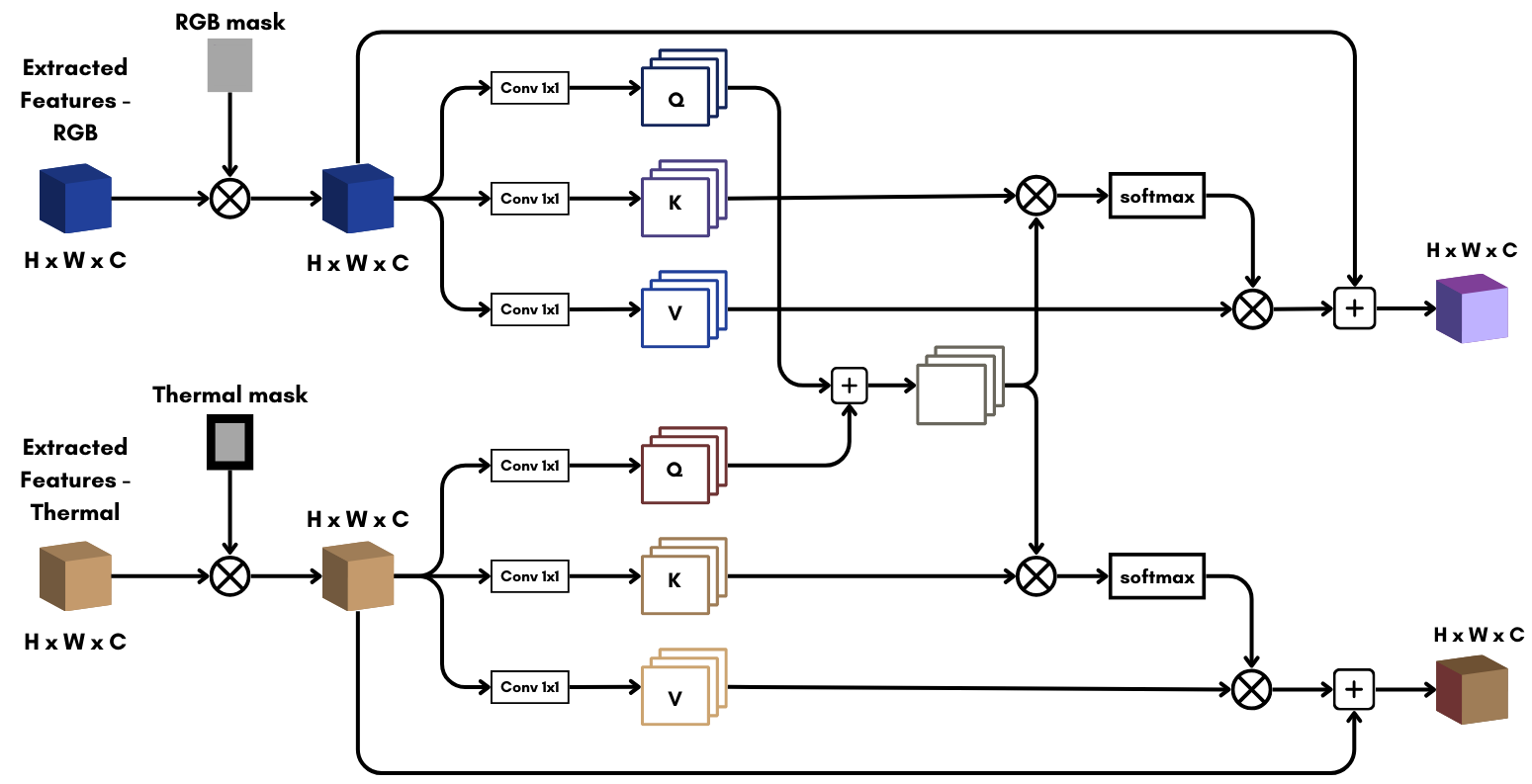}
  \vspace{-2mm}
  \caption{Illustration of the novel HA module developed. The use of masks to remove blackout regions helps in \textit{attending} to only regions with useful information. The results reported in Section \ref{experiments_results} show that the use of HA helps to reduce performance degradation during inference. Q,K,V depict Queries, Keys and Values.}
  \label{fig:HA}
  \vspace{-1em}
\end{figure*}

\begin{table}[b]
\centering
\caption{MobileNetV2~\cite{sandler2018mobilenetv2} architecture description.}
\vspace{-1mm}
\resizebox{\linewidth}{!}{%
\begin{tabular}{c|c|c|c|c|c|c}
\toprule
Input & Operator & \textit{t} & \textit{c} & \textit{n} & \textit{s} & Blocks                  \\ \toprule
\rowcolor[HTML]{EFEFEF} 
$224^2 \times 3$ & conv2d & -  & 32  & 1  & 2 &  \\
\rowcolor[HTML]{EFEFEF} 
$112^2 \times 32$ & bottleneck & 1  & 16  & 1  & 1 &  B1 \\  \rowcolor[HTML]{EFEFEF} 
$112^2 \times 16$ & bottleneck & 6  & 24  & 2  & 2 &  \\ \rowcolor[HTML]{EFEFEF} 
$56^2 \times 24$ & bottleneck & 6  & 32  & 3  & 2 &   \\ 
$28^2 \times 32$ & bottleneck & 6  & 64  & 4  & 2 & \multirow{2}{*}{B2} \\ 
$14^2 \times 64$ & bottleneck & 6  & 96  & 3  & 1 &  \\
\rowcolor[HTML]{EFEFEF} 
$14^2 \times 96$ & bottleneck & 6  & 160  & 3  & 2 & B3 \\
$7^2 \times 160$ & bottleneck & 6  & 320  & 1  & 1 & \multirow{4}{*}{N/A} \\
$7^2 \times 320$ & conv2d 1x1 & -  & 1280  & 1  & 1 &  \\
$7^2 \times 1280$ & avgpool 7x7 & -  & -  & 1  & - &  \\
$1 \times 1 \times 1280$ & conv2d 1x1 & -  & -  & - &   & - \\
\midrule
\multicolumn{7}{c}{\textit{Each line describes a sequence of 1 or more identical (modulo stride)}} \\
\multicolumn{7}{c}{\textit{layers, repeated $n$ times. All layers in the same sequence have the same}} \\
\multicolumn{7}{c}{\textit{number $c$ of output channels. The first layer of each sequence has a}}\\
\multicolumn{7}{c}{\textit{stride $s$ and all others use stride $1$; $t$ denotes the expansion factor~\cite{sandler2018mobilenetv2}.}} 
\end{tabular}}
\label{tab:mobilenetv2_architecture}
\vspace{-1em}
\end{table}

\subsection{HA-MLPD Overview}
Our model, HA-MLPD, consists of the feature extraction layers, HA module, fusion layer, and detection head. \Cref{fig:network_architecture} presents the general network architecture for the proposed HA-MLPD algorithm with the MobilenetV2 \cite{sandler2018mobilenetv2} backbone. Two separate branches are used to extract feature maps from both RGB and thermal images. The HA module, placed after the first block of layers, is used to perform cross- or self-attention accross the two modalities. The attention mechanism varies according to the overlapping and non-overlapping image regions (as illustrated in \Cref{fig:HA}). The extracted features from different levels are then concatenated and passed through a fusion layer to generate features used as input to the detection head. The fusion layer, designed to be as light as possible for real-time applications, is composed of one convolutional layer followed by batch normalization and ReLu activation. This is similar to MLPD, where this multi-level fusion strategy and fusion layer were introduced to cope with the loss of modality-specific information afer shared convolutional blocks. To strengthen this effect, we decoupled the modality-shared layers of MLPD to make them modality-specific in HA-MLPD. 

\subsection{Details of HA-MLPD Pipeline}

\textbf{Feature Extraction:} Any standard feature extraction backbone, possibly followed by additional blocks of layers, can be used to extract features ($F_{rgb}$, $F_{thermal}$) from both images:  
\begin{equation} \label{eq1}
\begin{split}
F_\mathrm{rgb} & = \phi_\mathrm{rgb} (I_\mathrm{rgb}) \ ,\\
F_\mathrm{thermal} & = \phi_\mathrm{thermal}(I_\mathrm{thermal}) \ ,
\end{split}
\end{equation}
\noindent where $I_\mathrm{rgb}$, $I_\mathrm{thermal}$ represents RGB and thermal images respectively, and $\phi_\mathrm{rgb}$, $\phi_\mathrm{thermal}$ the corresponding feature extractors. In our experiments, MobileNetV2~\cite{sandler2018mobilenetv2} was chosen for its compactness (see \Cref{fig:network_architecture} for overview of the corresponding architecture), and VGG-16~\cite{Simonyan15} as the original MLPD backbone. These feature extractors are composed of $B$ consecutive blocks of layers:
\begin{equation}
\begin{split}
\phi_\mathrm{rgb} &= \phi_\mathrm{rgb}^{(B)}\circ\cdots\circ\phi_\mathrm{rgb}^{(2)}\circ\phi_\mathrm{rgb}^{(1)} \ ,\\
\phi_\mathrm{thermal} &= \phi_\mathrm{thermal}^{(B)}\circ\cdots\circ\phi_\mathrm{thermal}^{(2)}\circ\phi_\mathrm{thermal}^{(1)} \ ,
\end{split}
\end{equation}

Features extracted after the first blocks $\phi_\mathrm{rgb}^{(1)}$, $\phi_\mathrm{thermal}^{(1)}$, denoted in what follows as $F_\mathrm{rgb}$, $F_\mathrm{thermal}$ for simplification, are then fed to our novel HA module.
\vspace{0.25em}

\textbf{Hybrid-Attention Module:} The HA module shown in \Cref{fig:HA} is the core of the proposed method. It enhances RGB and thermal features using the attention mechanism. The module switches between cross- and self-attention in response to the blackout regions in the input images. For that, we use masks $M_\mathrm{rgb}$, $M_\mathrm{thermal}$ corresponding to blackout regions to filter out features originating from these regions ($\otimes$ denotes element-wise multiplication): 
\begin{equation} \label{eq2}
\begin{split}
f_\mathrm{rgb} & = M_\mathrm{rgb}\otimes F_\mathrm{rgb} \ ,\\
f_\mathrm{thermal} & = M_\mathrm{thermal}\otimes F_\mathrm{thermal} \ .
\end{split}
\end{equation}

\noindent The Keys, Queries and Values for each modality are then generated using ${1\times1}$ convolution layers:
\begin{equation} \label{eq3}
\begin{split}
Q_\mathrm{rgb} & = \mathrm{Conv}_{1\times1}(f_\mathrm{rgb}),\\
K_\mathrm{rgb} & = \mathrm{Conv}_{1\times1}(f_\mathrm{rgb}),\\
V_\mathrm{rgb} & = \mathrm{Conv}_{1\times1}(f_\mathrm{rgb}),\\
Q_\mathrm{thermal} & = \mathrm{Conv}_{1\times1}(f_\mathrm{thermal}),\\
K_\mathrm{thermal} & = \mathrm{Conv}_{1\times1}(f_\mathrm{thermal}),\\
V_\mathrm{thermal} & = \mathrm{Conv}_{1\times1}(f_\mathrm{thermal}).
\end{split}
\end{equation}

\noindent The combined Query $Q_\mathrm{c}$ is then computed as the sum of the two modality-specific queries:
\begin{equation}
Q_\mathrm{c} = Q_\mathrm{rgb} + Q_\mathrm{thermal}.
\end{equation}
\noindent Removing the features corresponding to the blackout regions using the masks and then combining the queries will therefore make the modalities cross-attend where both modalities are available, and self-attend in the blackout regions. Following the standard practice, attended features $f^{*}_\mathrm{rgb}$ and $f^{*}_\mathrm{thermal}$ are calculated as: 
\begin{equation} \label{eq5}
\begin{split}
f^{*}_\mathrm{rgb} & = {\rm softmax}(Q^\top_\mathrm{c} K_\mathrm{rgb})V_\mathrm{rgb},\\
f^{*}_\mathrm{thermal} & = {\rm softmax}(Q^\top_\mathrm{c} K_\mathrm{thermal})V_\mathrm{thermal}.
\end{split}
\end{equation}

\noindent Finally, the enhanced features $f'_\mathrm{rgb}$ and $f'_\mathrm{thermal}$ for each modality are obtained as:
\begin{equation} \label{eq6}
\begin{split}
f'_\mathrm{rgb} & = f_\mathrm{rgb} + f^{*}_\mathrm{rgb},\\
f'_\mathrm{thermal} & =  f_\mathrm{thermal} + f^{*}_\mathrm{thermal}.
\end{split}
\end{equation}

\textbf{Fusion Layer:}
The enhanced features are then processed by the remaining feature extraction layers, to extract higher-level dual-modality-guided information. After each block of layers, the features from both branches are concatenated then fused using shared network layers. These layers are composed of 2D convolutions followed by Batch Norm and ReLu, similar to the MLPD baseline (see \Cref{fig:network_architecture}).

\vspace{0.5em}
\textbf{Detector Head:} The multi-level fused features are then passed through the detection head. In our architecture, and similar to MLPD, we use the SSD~\cite{SSDDetector} model for object detection. However, it can be replaced with any other state-of-the-art detector in practice. The detector outputs pedestrian bounding box locations and confidence scores.

\vspace{0.5em}
\textbf{Loss function:} The model is trained with regression loss on the bounding box locations $\mathscr{L}_{bbox}$ as in SSD \cite{SSDDetector} and multi-label loss $\mathscr{L}_{multilabel}$ from MLPD~\cite{kim2021mlpd}, balanced by a scaling factor $\lambda$:
\begin{equation}
\mathscr{L}=\mathscr{L}_{bbox}+\lambda \mathscr{L}_{multilabel}.
\end{equation}

\subsection{Masking Data Augmentation}
In addition to the data augmentations from the MLPD baseline, we included data augmentations using masking in our training process to foster the resilience of our approach. Our method involves masking the complete RGB and thermal modalities (probability of 10\% for each), as well as randomly masking patches of either modality again with a probability of $10\%$ for RGB masking and 10\% for thermal masking. Note that we avoid masking the same region in both modalities simultaneously. Also, these augmentations are implemented exclusively during the training phase and are not used during inference.

\section{Experiments and Results}
\label{experiments_results}

\begin{table*}[t]
\small
\centering
\caption{Experiment results on KAIST Dataset with two ({\normalfont ``Dual Modality"}) or only one ({\normalfont ``RGB Blackout", ``Thermal Blackout"}) modalities.}
\resizebox{0.9\textwidth}{!}{%
\begin{tabular}{l|ccc|ccc|ccc||c}
\rowcolor[HTML]{FFFFFF} 
\multicolumn{1}{c|}{} & \multicolumn{3}{c|}{Dual Modality} & \multicolumn{3}{c|}{RGB Blackout} & \multicolumn{3}{c||}{Thermal Blackout} & All \\

\multicolumn{1}{c|}{Method} & \begin{tabular}[c]{@{}c@{}}MR \\ (All) \end{tabular} & \begin{tabular}[c]{@{}c@{}}MR \\ (Day)\end{tabular} & \begin{tabular}[c]{@{}c@{}}MR \\ (Night)\end{tabular} & \begin{tabular}[c]{@{}c@{}}MR \\ (All)\end{tabular} & \begin{tabular}[c]{@{}c@{}}MR \\ (Day)\end{tabular} & \begin{tabular}[c]{@{}c@{}}MR \\ (Night)\end{tabular} & \begin{tabular}[c]{@{}c@{}}MR \\ (All)\end{tabular} & \begin{tabular}[c]{@{}c@{}}MR \\ (Day)\end{tabular} & \begin{tabular}[c]{@{}c@{}}MR \\ (Night)\end{tabular} & \begin{tabular}[c]{@{}c@{}}Average\\ difference \\\end{tabular} \\
\hline
\Tstrut\Bstrut
ACF~\cite{hwang2015multispectral} & 47.32 & 42.57 & 56.17 & {-} & {-} & {-} & {-} & {-} & {-} & {-} \\
Halfway Fusion~\cite{LiuZWM16} & 25.75 & 24.88 & 26.59 & {-} & {-} & {-} & {-} & {-} & {-} & {-} \\
FusionRPN+BF~\cite{KonigAJLNT17} & 18.29 & 19.57 & 16.27 & {-} & {-} & {-} & {-} & {-} & {-} & {-} \\
IAF R-CNN~\cite{LiSTT19} & 15.73 & 14.55 & 18.26 & {-} & {-} & {-} & {-} & {-} & {-} & {-} \\
IATDNN+IASS~\cite{GuanCYCY19} & 14.95 & 14.67 & 15.72 & {-} & {-} & {-} & {-} & {-} & {-} & {-} \\
CIAN~\cite{ZhangLZYQHH19} & 14.12 & 14.77 & 11.13 & {-} & {-} & {-} & {-} & {-} & {-} & {-} \\
MSDS-RCNN~\cite{li_2018_BMVC} & 11.34 & 10.53 & 12.94 & 36.36 & 39.53 & 28.67 & 82.97 & 76.04 & 97.68 & +27.60\\
AR-CNN~\cite{ZhangZC0LL19} & 9.34 & 9.94 & 8.38 & 17.70 & 21.95 & 8.64 & 77.03 & 67.54 & 97.85 & +18.73\\
MBNet~\cite{zhou2020improving} & 8.13 & 8.28 & 7.86 & 55.56 & 57.49 & 46.81 & 80.20 & 71.88 & 100 & +32.01\\
SSD-RGB~\cite{SSDDetector} & {-} & {-} & {-} & {-} & {-} & {-} & 34.63 & 25.38 & 53.86 & {-} \\
SSD-Thermal~\cite{SSDDetector} & {-} & {-} & {-} & 21.12 & 25.63 & 12.58 & {-} & {-} & {-} & {-} \\
MLPD~\cite{kim2021mlpd} & \textcolor{black}{7.58} & \textcolor{black}{7.95} & \textcolor{black}{6.95} & \textcolor{black}{16.34} & \textcolor{blue}{\it 20.07} & \textcolor{black}{8.22} & \textcolor{blue}{\it 23.95} & \textcolor{blue}{\it 16.88} & \textcolor{blue}{\it 39.37} & 0\\ 
\hline 
\Tstrut\Bstrut
HA-MLPD (VGG-16) & \textcolor{blue}{\it 5.19} & \textcolor{blue}{\it 6.16} & \textcolor{blue}{\it 3.81} & \textcolor{blue}{\it 15.21} & \textcolor{black}{20.41} & \textcolor{blue}{\it 4.88} & \textcolor{red}{\bf 20.6} & \textcolor{red}{\bf 15.86} & \textcolor{red}{\bf 30.76} & \textcolor{red}{\bf -2.29}\\ 
HA-MLPD (MobileNetv2) & \textcolor{red}{\bf 4.42} & \textcolor{red}{\bf 5.42} & \textcolor{red}{\bf 2.55} & \textcolor{red}{\bf 13.9} & \textcolor{red}{\bf 18.38} & \textcolor{red}{\bf 3.62} & \textcolor{black}{24.29} & \textcolor{black}{16.92} & \textcolor{black}{40.83} & \textcolor{blue}{\it -1.75}\\
\end{tabular}}
\label{tab:res-dual-failure}
\end{table*}

\begin{table*}[h]
\small
\centering
\caption{Experiment results on KAIST Dataset with partial overlap between modalities.}
\resizebox{0.9\textwidth}{!}{%
\begin{tabular}{l|ccc|ccc|ccc||c}
\multicolumn{1}{c|}{} & \multicolumn{3}{c|}{Sides Blackout (RGB-Thermal)} & \multicolumn{3}{c|}{Sides Blackout (Thermal-RGB)} & \multicolumn{3}{c||}{Surrounding Blackout} & All \\
\multicolumn{1}{c|}{Method} & \begin{tabular}[c]{@{}c@{}}MR \\ (All)\end{tabular} & \begin{tabular}[c]{@{}c@{}}MR \\ (Day)\end{tabular} & \begin{tabular}[c]{@{}c@{}}MR \\ (Night)\end{tabular} & \begin{tabular}[c]{@{}c@{}}MR \\ (All)\end{tabular} & \begin{tabular}[c]{@{}c@{}}MR \\ (Day)\end{tabular} & \begin{tabular}[c]{@{}c@{}}MR \\ (Night)\end{tabular} & \begin{tabular}[c]{@{}c@{}}MR \\ (All)\end{tabular} & \begin{tabular}[c]{@{}c@{}}MR \\ (Day)\end{tabular} & \begin{tabular}[c]{@{}c@{}}MR \\ (Night)\end{tabular} & \begin{tabular}[c]{@{}c@{}}Avg.\\ diff. \\\end{tabular} \\
\hline
\Tstrut\Bstrut
MSDS-RCNN~\cite{li_2018_BMVC} & 43.00 & {-} & {-} & 59.42 & {-} & {-} & 47.22 & {-} & {-} & +30.05\\
AR-CNN~\cite{ZhangZC0LL19} & 32.18 & {-} & {-} & 54.59 & {-} & {-} & 57.58 & {-} & {-} & +28.28\\
MBNet~\cite{zhou2020improving} & 56.65 & {-} & {-} & 63.81 & {-} & {-} & 46.99 & {-} & {-} & +35.98\\
SSD-RGB~\cite{SSDDetector} & 63.75 & {-} & {-} & 51.08 & {-} & {-} & 34.63 & {-} & {-} & +29.99\\
SSD-Thermal~\cite{SSDDetector} & 38.73 & {-} & {-} & 59.98 & {-} & {-} & 55.06 & {-} & {-} & +31.42\\
MLPD*~\cite{kim2021mlpd} & \textcolor{black}{17.31} & \textcolor{black}{19.46} & \textcolor{black}{13.44} & \textcolor{black}{23.03} & \textcolor{black}{20.41} & \textcolor{black}{27.38} & \textcolor{black}{19.16} & \textcolor{black}{18.11} & \textcolor{black}{22.41} & 0\\
\hline
\Tstrut\Bstrut
HA-MLPD (VGG-16) & \textcolor{blue}{\it 13.40} & \textcolor{blue}{\it 15.32} & \textcolor{red}{\bf 9.55} & \textcolor{red}{\bf 18.24} & \textcolor{red}{\bf 18.11} & \textcolor{red}{\bf 18.97} & \textcolor{blue}{\it 11.84} & \textcolor{red}{\bf 10.94} & \textcolor{blue}{\it 14.43} & \textcolor{blue}{\it -5.34}\\
HA-MLPD (MobileNetv2) & \textcolor{red}{\bf 12.84} & \textcolor{red}{\bf 13.90} & \textcolor{blue}{\it 10.34} & \textcolor{blue}{\it 19.29} & \textcolor{blue}{\it 18.67} & \textcolor{blue}{\it 20.97} & \textcolor{red}{\bf 11.26} & \textcolor{blue}{\it 11.56} & \textcolor{red}{\bf 11.25} & \textcolor{red}{\bf -5.37}\\ 
\multicolumn{11}{c}{ \textit{*: MR values for day and night images were computed from pre-trained model released by authors; not reported in \cite{kim2021mlpd}.}}
\end{tabular}}
\label{tab:res-unpaired}
\vspace{-0.5em}
\end{table*}

\subsection{Dataset}

The KAIST Multispectral Pedestrian Dataset \cite{hwang2015multispectral} consists of 95,328 RGB-Thermal pairs fully overlapped captured in an urban environment. The provided ground truth consists of 103,128 pedestrian bounding boxes in 1,182 instances. In our experiments, we sample 1 frame out of 2 so that 25,076 frames are used for training, as in \cite{kim2021mlpd}. For evaluation, we also follow the standard evaluation criterion, which consists of sampling 1 out of every 20 frames, so the results are evaluated in 2,252 frames where 1,455 frames were recorded during the day and 797 frames at night \cite{kim2021mlpd}. We use paired annotations for training \cite{ZhangZC0LL19} and sanitised annotations for evaluation \cite{li_2018_BMVC}.

To generate the complete blackout cases arising from sensor failures, the original pixels values are replaced with zeros for either of the modalities as shown in \Cref{fig:blackout_scenarios}. For sides blackout cases, the original images are divided into three equal-sized vertical portions, and opposite side portions (\textit{e.g.}, left part in the RGB image and right part in the thermal image) are replaced with zero value pixels in each modality. For the surrounding blackout scenarios, we centre-crop the thermal image (96 pixels cropped at both the top and bottom and 120 pixels on the left and right sides) and replace the removed regions with zero pixels, while retaining the entire RGB image. 

\subsection{Training details}

\textbf{HA-MLPD with MobileNetV2 backbone --} We extended the MobileNetV2 architecture with further convolutional layers, as shown in \Cref{fig:network_architecture}. To initialise, we used ImageNet pre-trained weights for blocks (B1, B2, B3) and the remaining convolution kernels were initialised using values sampled from a normal distribution (std=0.01). The network training process spanned 200 epochs, with an early stop callback in place to stop training if no improvement was observed for 50 epochs. The model was trained using SGD with an initial LR, momentum, and weight decay set at $1e^{-3}$, 0.9, and $4e^{-5}$, respectively. The LR was scheduled to decrease at the 150th and 190th epochs with a gamma value of 0.1.

\vspace{1em}
\textbf{HA-MLPD with VGG-16 backbone --} Like the original MLPD, we use VGG16 pre-trained on ImageNet with batch normalisation, from $Conv1$ to $Conv5$, and the remaining convolution kernels are initialised with values drawn from the normal distribution (std=0.01). The HA module is adopted at the output of $Conv4$ \cite{kim2021mlpd}. The model is trained by Stochastic Gradient Descent (SGD) with the initial learning rate (LR), momentum, and weight decay, as $1e^{-4}, 0.9,$ and $5e^{-4}$, respectively. The mini-batch size is set to 8 and the input image size is resized to 512 (H) x 640 (W). The network was trained for 40 epochs, and the LR was scheduled to decrease after 20 and 36 epochs. The models were trained on a single GPU accelerator node featuring 2x AMD Rome CPUs (32 cores @ 2.35 GHz) and 4x NVIDIA A100-40 GPUs.

\subsection{Adverse inference-time conditions simulation}
To simulate sensor failure, we replaced one image of the RGB-Thermal pair by a full black image (zero pixel values). To simulate partial overlapping, we replace different parts of the images by black regions (see \Cref{fig:blackout_scenarios}). Side blackouts simulate discrepancies in lateral fields of view, whereas surrounding blackouts simulate differences in either focal length or sensor resolution.

\subsection{Metrics}
\begin{figure*}
    \centering
    \includegraphics[trim=5.1cm 5.5cm 2.9cm 5.7cm, clip, width=\linewidth]{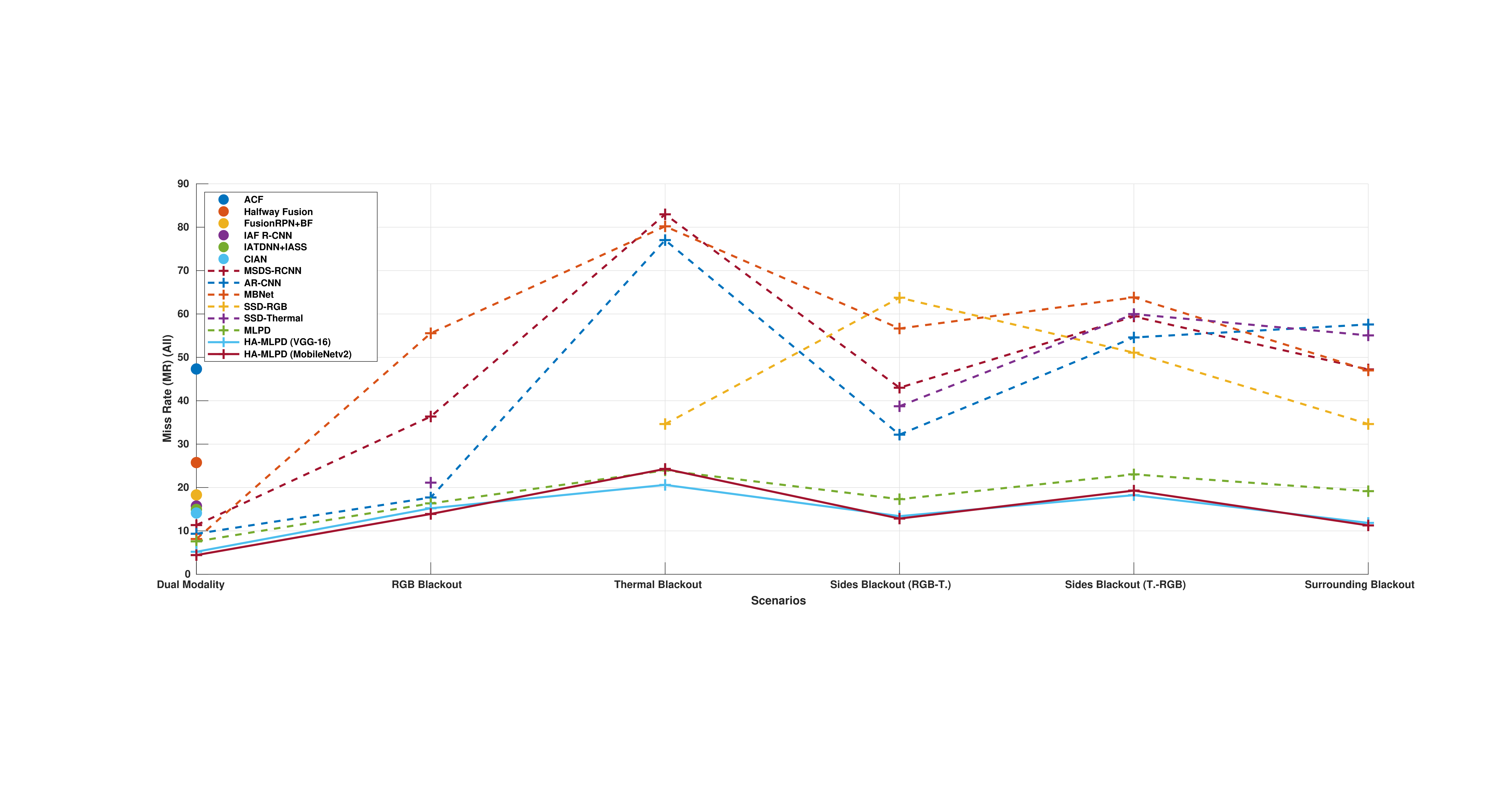}
    \caption{Visualization of experiment results on KAIST Dataset.}
    \label{fig:results}
\end{figure*}

\begin{table*}[t]
\small
\centering
\caption{Ablation study and comparison with the {\normalfont MLPD} baseline.}
\resizebox{0.9\textwidth}{!}{%
\begin{tabular}{l|ccc|ccc|ccc||c}
\multicolumn{1}{c|}{} & \multicolumn{3}{c|}{Sides Blackout (RGB-Thermal)} & \multicolumn{3}{c|}{Sides Blackout (Thermal-RGB)} & \multicolumn{3}{c||}{Surrounding Blackout} & All \\

\multicolumn{1}{c|}{Method} & \begin{tabular}[c]{@{}c@{}}MR \\ (All)\end{tabular} & \begin{tabular}[c]{@{}c@{}}MR \\ (Day)\end{tabular} & \begin{tabular}[c]{@{}c@{}}MR \\ (Night)\end{tabular} & \begin{tabular}[c]{@{}c@{}}MR \\ (All)\end{tabular} & \begin{tabular}[c]{@{}c@{}}MR \\ (Day)\end{tabular} & \begin{tabular}[c]{@{}c@{}}MR \\ (Night)\end{tabular} & \begin{tabular}[c]{@{}c@{}}MR \\ (All)\end{tabular} & \begin{tabular}[c]{@{}c@{}}MR \\ (Day)\end{tabular} & \begin{tabular}[c]{@{}c@{}}MR \\ (Night)\end{tabular} & \begin{tabular}[c]{@{}c@{}}Average\\ difference \\\end{tabular} \\
\hline
\Tstrut\Bstrut
MLPD \cite{kim2021mlpd} & 17.31 & 19.46 & 13.44 & 23.03 & 20.41 & 27.38 & 19.16 & 18.11 & 22.41 & 0 \\
HA-MLPD w/o HA & \textcolor{blue}{\it 15.66} & \textcolor{blue}{\it 17.81} & \textcolor{blue}{\it 10.65} & \textcolor{blue}{\it 19.79} & \textcolor{blue}{\it 19.64} & \textcolor{blue}{\it 20.6} & \textcolor{blue}{\it 12.48} & \textcolor{blue}{\it 13.17} & \textcolor{red}{\bf 10.63} & \textcolor{blue}{\it -3.86}\\
HA-MLPD & \textcolor{red}{\bf 13.40} & \textcolor{red}{\bf 15.32} & \textcolor{red}{\bf 9.55} & \textcolor{red}{\bf 18.24} & \textcolor{red}{\bf 18.11} & \textcolor{red}{\bf 18.97} & \textcolor{red}{\bf 11.84} & \textcolor{red}{\bf 10.94} & \textcolor{blue}{\it 14.43} & \textcolor{red}{\bf -5.34}\\
\end{tabular}}
\label{tab:ablation-augm}
\vspace{-0.5em}
\end{table*}

Following the standard practice in the field, and especially the MLPD baseline~\cite{kim2021mlpd}, we report the log-average Miss Rate (MR)~\cite{5975165} at an Intersection-over-Union threshold of 0.5 to summarise the detector performance. This metric gives a stable and informative assessment of detector performance~\cite{5975165}. 

\subsection{Results}
Table~\ref{tab:res-dual-failure} presents a comparison of the performance of our method with the existing literature in pedestrian detection scenarios using dual RGB and thermal modalities, especially in cases of sensor failure, such as RGB and thermal blackouts. Our method consistently shows competitive or superior results, demonstrating robustness in blackout conditions, particularly excelling in scenarios with RGB blackouts. It is important to highlight that even in complete blackouts of one modality, our method, in which HA then collapses in self-attention, outperforms both the RGB-only and Thermal-only models (referred to as SSD-RGB \& SSD-Thermal). Furthermore, Table~\ref{tab:res-dual-failure} includes comparisons with MLPD~\cite{kim2021mlpd}. The colour and font coding highlights the best and second-best results in \textcolor{red}{\bf bold red} and \textcolor{blue}{\it italic blue}, respectively. HA-MLPD with the MobileNetv2 backbone achieves the highest performance in dual modality and RGB blackout scenarios, while HA-MLPD with the original VGG-16 backbone excels when RGB information is unavailable. In summary, our method demonstrates state-of-the-art performance, underscoring the effective RGB-Thermal fusion strategy in managing diverse real-world inference conditions.

Table~\ref{tab:res-unpaired} delves into the performance of the different methods in the context of inference-time misalignments involving extrinsic and intrinsic discrepancies in cameras. The scenarios considered include Sides Blackout with RGB-Thermal and Thermal-RGB misalignments, as well as Surrounding Blackout; see Figure~\ref{fig:blackout_scenarios}. Our method consistently exhibits the best performance across these misalignment scenarios, outperforming baseline methods such as MSDS-RCNN~\cite{li_2018_BMVC}, AR-CNN~\cite{ZhangZC0LL19}, MBNet~\cite{zhou2020improving}, SSD-RGB~\cite{SSDDetector}, SSD-Thermal~\cite{SSDDetector} and MLPD~\cite{kim2021mlpd}. 
Both backbones lead to second or best performance, with an average margin of more than 5 percentage points ($pp$) with respect to competitors.
Overall, the table emphasises the robustness of HA-MLPD to inference-time blackouts, showcasing its efficacy in scenarios encountered in real-world applications. Fig.~\ref{fig:results} summarizes the results from Tables~\ref{tab:res-dual-failure} and \ref{tab:res-unpaired} in a more easily interpretable plot.

\subsection{Ablation Study}

The ablation study in Table~\ref{tab:ablation-augm} investigates the impact of our contributions in inference-time blackout scenarios, specifically Sides Blackout with RGB-Thermal, Sides Blackout with Thermal-RGB and surrounding blackout conditions. The table outlines the MR results for different configurations: (1) MLPD, (2) HA-MLPD without the HA mechanism (w/o HA) but with masking data augmentation, and (3) our proposed HA-MLPD. All three methods are with the VGG-16 backbone. The results demonstrate a systematic improvement with each modification, with lower MR values indicating better performance. The addition of data augmentation (Aug) results in a notable reduction in MR values across all scenarios ($-3.86pp$ on average). Furthermore, incorporation of HA contributes to an additional improvement, achieving the lowest MR values ($-5.34pp$ on average, compared to MLPD). Overall, the ablation study underscores the significance of both HA and data augmentation in improving the model's robustness under inference-time blackouts.

\subsection{Discussion}

\paragraph{Dual modality performance} The results achieved in the dual modality scenario, \textit{i.e.} when both modalities are fully overlapping and available, placed HA-MLPD in the second place (username: UniLu) on the leaderboard\footnote{\href{https://eval.ai/web/challenges/challenge-page/1247/leaderboard/3137}{https://eval.ai/web/challenges/challenge-page/1247/leaderboard/3137}} at the time of submission. It is worth noting that this places our method above competitors that were specifically designed for dual modality conditions only, whence not robust to blackout scenarios. On the contrary, our method operates a necessary trade-off to cope with such challenging conditions, but still outperforms most of other methods under the normal conditions.

\paragraph{Model efficiency} In our evaluation, we used two models, one with the VGG-16 backbone, which contains a total of 59.88 M parameters, and the other with the MobileNetV2 backbone, which contains a total of 13.96 M with a 4x reduction in the total number of parameters. Due to its larger size, the VGG-16 backbone required more time to train per epoch with a smaller batch size, while the smaller MobileNetV2 backbone allowed faster iteration through each epoch. However, MobileNetV2 required training for more number of epochs to achieve the best results. The overall duration of the training slightly favoured MobileNetV2. Given that the MobileNetV2 backbone is smaller, it can be easily accommodated on hardware with limited resources and can deliver a higher frame-per-second (FPS) rate during inference without compromising the performance. Note that our model takes approx. 20ms (compared to 42ms for MLPD baseline) to process an image during inference on a desktop workstation containing an Intel i7-11700 @ 2.50GHz CPU and 1x NVIDIA GeForce RTX 3090 GPU.

\paragraph{Reliance on masks} Our approach relies on the masks of the blackout regions to guide the network through hybrid-attention mechanisms. The success of the proposed method depends on the accuracy of these masks. If the blackout regions are not correctly identified, the HA mechanism could be misguided, leading to suboptimal pedestrian detection performance.

\section{Conclusion}
\label{section:conclusion_and_future_works}
In this paper, we tackle the challenges of multispectral pedestrian detection in real-world scenarios, such as adverse inference-time configurations and hardware resource limitations. We introduced a novel HA mechanism to mitigate performance degradation arising from modality-specific lack of information. During training, the HA module enables learning both generalised and discriminative features through self- or cross-attention mechanisms. A mobile-friendly backbone is also used for higher energy efficiency. Extensive experimental comparisons demonstrate that the proposed method is robust to realistic stereo conditions. Furthermore, an ablation study highlighted that both introduced HA mechanism and data augmentation play crucial roles in improving the model's robustness. Future work will include deploying the method on edge device for real-condition testing, and evaluating the HA mechanism on other modalities (\textit{e.g.}, depth images).

\section*{Acknowledgment}
The simulations were performed on the Luxembourg National Supercomputer MeluXina. The authors gratefully acknowledge the LuxProvide teams for their expert support.

\ifCLASSOPTIONcaptionsoff
  \newpage
\fi

{\small
\bibliographystyle{IEEEtran}
\bibliography{main}
}

\end{document}